\documentclass[runningheads]{llncs}

\usepackage{eccv}

\usepackage{eccvabbrv}

\usepackage{graphicx}
\usepackage{booktabs}
\usepackage{bm}
\usepackage{multirow}
\usepackage{colortbl} 
\usepackage{amsmath,amssymb}
\usepackage{algorithm}
\usepackage{algpseudocode}
\usepackage{wrapfig}

\usepackage[accsupp]{axessibility}  

\usepackage{hyperref}

\usepackage{orcidlink}

\begin{document}


\title{Any Target Can be Offense: Adversarial Example Generation via Generalized Latent Infection} 

\titlerunning{Adversarial Example Generation via Generalized Latent Infection}


\author{Youheng Sun\inst{1}\inst{\star}\orcidlink{0009-0006-3110-1658} \and
Shengming Yuan\inst{1}\thanks{Equal contribution.}\orcidlink{0009-0003-0183-2976}\and
Xuanhan Wang\inst{2}\thanks{Corresponding author}\orcidlink{0000-0002-3881-9658} \and
Lianli Gao\inst{1}\orcidlink{0000-0002-2522-6394} \and
Jingkuan Song\inst{1}\orcidlink{0000-0002-2549-8322}
}

\authorrunning{Y.Sun et al.}

\institute{Center for Future Media, University of Electronic Science and Technology of China
\and
Shenzhen Institute for Advanced Study, University of Electronic Science and Technology of China\\
\email{youheng.sun@std.uestc.edu.cn, shengming.yuan@outlook.com, 
wxuanhan@hotmail.com,
lianli.gao@uestc.edu.cn, jingkuan.song@gmail.com}
}

\maketitle

\begin{abstract}

Targeted adversarial attack, which aims to mislead a model to recognize any image as a target object by imperceptible perturbations, has become a mainstream tool for vulnerability assessment of deep neural networks (DNNs).  
Since existing targeted attackers only learn to attack known target classes, they cannot generalize well to unknown classes.
To tackle this issue, we propose \textbf{G}eneralized \textbf{A}dversarial attac\textbf{KER} (\textbf{GAKer}), which is able to construct adversarial examples to any target class. The core idea behind GAKer is to craft a latently infected representation during adversarial example generation. 
To this end, the extracted latent representations of the target object are first injected into intermediate features of an input image in an adversarial generator. 
Then, the generator is optimized to ensure visual consistency with the input image while being close to the target object in the feature space.
Since the GAKer is class-agnostic yet model-agnostic, it can be regarded as a general tool that not only reveals the vulnerability of more DNNs but also identifies deficiencies of DNNs in a wider range of classes.  
Extensive experiments have demonstrated the effectiveness of our proposed method in generating adversarial examples for both known and unknown classes. 
Notably, compared with other generative methods, our method achieves an approximately \textbf{14.13\%} higher attack success rate for unknown classes and an approximately \textbf{4.23\%} higher success rate for known classes. Our code is available in \url{https://github.com/VL-Group/GAKer}.

  \keywords{Targeted Adversarial Attack \and Generator-based Attack \and Black-box attack \and Unknown Classes}
\end{abstract}

\section{Introduction}
\label{sec:intro}

Deep neural networks (DNNs) have significantly advanced the field of artificial intelligence, achieving remarkable success in various domains, including image recognition~\cite{he2016deep}, natural language processing~\cite{ashish2017attention_is, hugo2023llama}, and 
AIGC~\cite{NEURIPS2020_4c5bcfec,su2024f3}. 
Despite great success, DNNs have been shown to be significantly vulnerable to adversarial attacks~\cite{szegedy2014intriguing, goodfellow2015explainingFGSM, naseer2022on,gu2023survey_atrans}, which misleads DNNs to fail by using adversarial examples, i.e., adding human-imperceptible perturbations into clean images. Thus, it is of great importance to understand the mechanism behind DNNs and design effective assessment methods~\cite{szegedy2014intriguing, goodfellow2015explainingFGSM, naseer2022on,gu2023survey_atrans, yuan2022NCF} to identify deficiencies of DNNs before deploying them in security-sensitive applications.

In terms of the way of attacking, adversarial attacks generally can be divided into two categories. The first one is the untargeted attack~\cite{goodfellow2015explainingFGSM, IFGSM,dong2018boosting_MIFGSM,dong2019evading_TIFGSM,xie2019improving_DIFGSM, yuan2022NCF}, where the goal of attackers is to fail DNNs. The second one is the targeted attack~\cite{zhao2023minimizing, yang2022HGN, gao2023ESMA, naseer2021TTP}, where attackers not only fail DNNs but also mislead them to recognize an image as the pre-specific target. Since the targeted attack with high flexibility poses a severe threat to security-sensitive applications, it has become a mainstream tool for vulnerability assessment of DNNs. Therefore, in this work, we focus on the study of targeted attacks.

\begin{figure}[tb]
  \centering
  \begin{subfigure}{0.3\linewidth}
    \includegraphics[width=\linewidth]{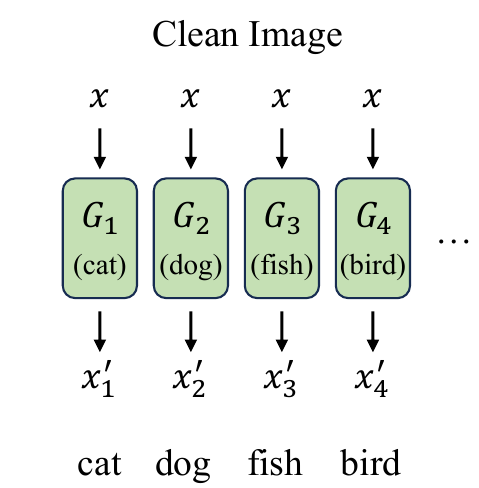}
    \caption{Single-target attack}
    \label{fig:single-target}
  \end{subfigure}
  \hfill
  \begin{subfigure}{0.3\linewidth}
    \includegraphics[width=\linewidth]{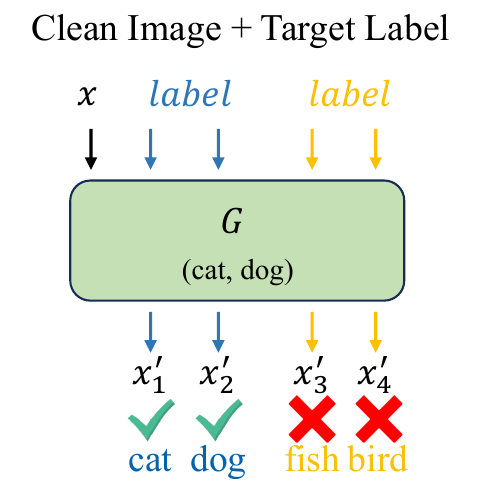}
    \caption{Multiple-target attack}
    \label{fig:multi-target}
  \end{subfigure}
  \hfill
  \begin{subfigure}{0.3\linewidth}
    \includegraphics[width=\linewidth]{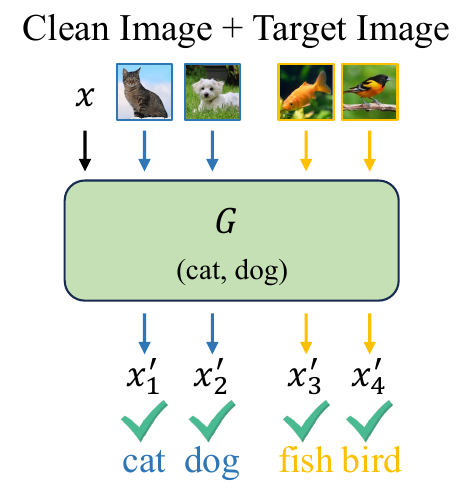}
    \caption{Arbitrary-target attack}
    \label{fig:any-target}
  \end{subfigure}
  \caption{\textbf{The inference process of different generator-based targeted attacks. }
In the center of each scenario is the generator $ \mathcal{G}$ that takes the model inputs at the top and aims to produce adversarial examples $\bm{x'}$ at the bottom. The classes indicated within $\mathcal{G}$ (cat, dog, fish, bird) are the training classes. \textcolor{blue}{Blue lines} denote known classes that are encountered during inference, and \textcolor{orange}{yellow lines} denote unknown classes that were not present in the training data. Sub-figure \textbf{(a)} depicts a single-target attack where each generator is specialized for one class, thus can only attack that specific class. Sub-figure \textbf{(b)} demonstrates a multiple-target attack where the generator $\mathcal{G}$ takes a source image $\bm{x}$ and known target labels (\eg, cat, dog) to create their adversarial examples $\bm{x'}$, but it fails to attack labels unknown to the training (\eg, fish, bird). Sub-figure \textbf{(c)} represents an arbitrary-target attack where $\mathcal{G}$ can utilize target images to craft adversarial examples capable of misleading the classifier into known and unknown classes (\eg, fish, bird), highlighting the generalization capability of this approach.
  }
  \label{fig:paradigm}
\end{figure}

Among the target attack methods, two technical branches exist for adversarial example generation. The first one is the iteration-based framework, which produces an adversarial example of each clean image in an iterative manner. This framework has shown to be susceptible to overfitting white-box models, and iteration-based strategy leads to a heavy computational overhead~\cite{zhao2023minimizing, gao2023ESMA, yang2022HGN, han2019MAN}. In a different line of this, the second branch is the generator-based framework, which constructs an adversarial example by using a trained generative model and shows a great potential of transferability~\cite{zhao2023minimizing, gao2023ESMA, yang2022HGN, han2019MAN}. However, the generator used in existing methods~\cite{naseer2021TTP,poursaeed2018Omid_GAP,naseer2021on,wang2023TTAA,zhao2023minimizing} is trained to adapt adversarial examples to known target classes only. As depicted in \cref{fig:single-target} and \cref{fig:multi-target}, the trained generator is responsible for either only one class or a set of known classes. When it comes to an unknown target class (\ie, the class not seen during training), previous methods are not capable of generating relatively adversarial examples unless retraining the generator, thus limiting the comprehensive assessment of DNNs. To tackle this, one straightforward solution is to train a generator to adapt a vast number of target classes with a large-scale dataset (\eg, ImageNet-21k). However, as demonstrated in~\cite{yang2022HGN}, the attack success rate of an adversarial example significantly degrades as the number of known classes increases. Hence, one question arises: \textit{how to design a generalized yet efficient assessment in which vulnerability of DNNs can be evaluated by any target classes?}

To answer the above question, we study a more practical paradigm. As shown in \cref{fig:any-target}, any target object could be a good offense, and an adversarial example can be constructed from any target regardless of whether it is a known class or not. To achieve such generalization capability, we argue that extracting the major component of an object is the key to adversarial example generation. Motivated by this, we propose \textbf{G}eneralized \textbf{A}dversarial attac\textbf{KER}, termed (\textbf{GAKer}). 
The core idea behind the GAKer is to contaminate the latent representation of a clean image with the major component of the target object. To equip GAKer with the capability of latent infection, it jointly utilizes the latent representation of a clean image and the major component of the target object to generate the adversarial example. Then, the adversarial example generated from GAKer is optimized to remain visually consistent with the clean image, but the corresponding major component is dominated by the target object. 
Once trained, the GAKer has the ability to replace major components between two images without depending on specific class or targeted DNN, thus improving the generalization capability of DNN assessment.
Comprehensive evaluations across diverse DNNs, encompassing standard models, adversarially trained models, and vision-language foundation models, reveal that the proposed GAKer can effectively generate high-quality adversarial examples regardless of target classes. This demonstrates GAKer's generalization ability, making it a valuable tool for the adversarial robustness assessment of DNNs.
In summary, three contributions are highlighted:
\begin{itemize}
    
    \item We propose a novel \textbf{G}eneralized \textbf{A}dversarial attac\textbf{KER}, which is a general assessment tool since it can generate adversarial examples from any object. Without satisfying the visual appearance of an image, it can mislead any DNNs by latently changing the major components of an image.
    \item  To our knowledge, this work, for the first time, explores the problem of generalized target adversarial attack. Our study reveals that changing the major components of an object is the key to the generalized assessment of DNNs. 
    \item Extensive experiments conducted on a wide range of DNNs demonstrate the generalizability of our proposed method for vulnerability assessment, especially under the setting of any targeted class. 
    Particularly, our method has increased the targeted attack success rate by approximately 14.13\% for unknown classes and by approximately 4.23\% for known classes compared with other generator-based approaches.
\end{itemize}

\section{Related Work}

\subsection{Iterative Methods}
\label{sec: iterative-based}

Since the discovery of adversarial examples, most iterative methods are proposed, which utilize model gradients to iteratively add adversarial perturbations to specified images. These methods are mainly categorized as gradient-based optimization and input transformation.
The gradient-based optimization aims to circumvent poor local optima by employing optimization techniques. MI-FGSM ~\cite{dong2018boosting_MIFGSM} and NI-FGSM~\cite{Lin2020Nesterov_NI} introduce momentum and Nesterov accelerated gradient into the iterative attack process to enhance black-box transferability, respectively. PI-FGSM~\cite{gao2020patch} introduces patch-wise perturbations to better cover the discriminative region. VMI-FGSM~\cite{Wang2021enhancing_VMI} tunes the current gradient with the gradient variance from the neighborhood. RAP~\cite{qin2022RAP} advocates injecting worst-case perturbations at each step of the optimization procedure rather than minimizing the loss of individual adversarial points. The input transformation methods also increase adversarial transferability by preventing overfitting to the surrogate model. DI-FGSM~\cite{xie2019improving_DIFGSM} applies various input transformations to the clean images. 
SIT~\cite{wang2023SIT} applies a random image transformation onto each image block to generate a diverse set of images for gradient calculation. SU~\cite{wei2023SU} introduces a feature similarity loss to encourage universal learned perturbations by maximizing the similarity between the global adversarial perturbation and randomly cropped local regions.

\subsection{Generative Methods}
\label{sec: generative-based}

Another branch of targeted attacks utilizes generators to craft adversarial examples. 
Compared with iterative-based attacks, generator-based attacks have several characteristics ~\cite{gu2023survey_atrans}: high efficiency with just a single model-forward pass at test time and superior generalizability through learning the target distribution rather than class-boundary information~\cite{naseer2021TTP}. 
Thus, many generator-based targeted attack methods are proposed, which can divided into single-target and multi-target generator attacks. 
Notably, this work focuses on scenarios where black-box models are entirely inaccessible, so query-based generator attacks which requiring extensive querying are not within the scope of discussion.

\textbf{Single-target Generative Methods:}
Early generative targeted attacks employed a single generator to attack a specific target, primarily aiming to enhance transferability across various models. 
Pourseed~\cite{poursaeed2018Omid_GAP} proposes a generator capable of producing image diagnostic and image-dependent perturbations for targeted attacks. Naseer~\cite{naseer2019CDAP} introduced a relativistic training objective to mislead networks trained on completely different domains. Furthermore, Naseer~\cite{naseer2021TTP} matches the perturbed image distribution with that of the target class. TTAA~\cite{wang2023TTAA} captures the distribution information of the target class from both label-wise and feature-wise perspectives to generate highly transferable targeted adversarial examples.

\textbf{Multi-target Generative Methods:}
However, generator-based methods are confined to single-target attacks, which require training a generative model for each target class, resulting in considerable computational costs and inefficiency.
Recently, exemplified by the introduction of the Multi-target Adversarial Network (MAN) framework~\cite{han2019MAN}, have revolutionized this landscape. MAN represents a paradigm shift by enabling the generation of adversarial examples across multiple target classes through a unified training process. Yang \etal~\cite{yang2022HGN} introduce a significant contribution by leveraging a hierarchical generative network. Through this design, they are able to train 20 generators, each trained for 50 target classes, thereby covering all 1000 classes in the ImageNet dataset. Gao \etal~\cite{gao2023ESMA} proposes a generative targeted attack strategy named Easy Sample Matching Attack (ESMA), which exhibits a higher success rate for targeted attacks through generating perturbations towards High-Sample-Density-Regions of the target class.

\subsection{Adversarial Defenses}
\label{sec:defend}

A primary class of defense methods processes adversarial images to break the perturbations. For instance, Guo \etal~\cite{guo2018countering} introduces several techniques for input transformation, such as JPEG compression~\cite{guo2018countering}, to mitigate adversarial perturbations. R\&P~\cite{xie2018mitigating} employs random resizing and padding to reduce adversarial effects. HGD~\cite{liao2018defense} develops a high-level representation guided denoiser to diminish the impact of adversarial disturbances. ComDefend~\cite{jia2019comdefend} proposes an end-to-end image compression model to defend against adversarial examples. NRP~\cite{naseer2020a} trains a neural representation purifier model that removes adversarial perturbations using automatically derived supervision.
Another approach enhances resilience to attacks by incorporating adversarial examples into the training phase. For instance, Tramèr \etal~\cite{tramer2018ensemble} bolster black-box robustness by utilizing adversarial examples generated from unrelated models. Similarly, Xie \etal~\cite{xie2019feature} integrate feature denoising modules trained on adversarial examples to develop robustness in the white-box model.

\section{Methodology}

\subsection{Problem Formulation}
\label{sec:problem}

Formally, let $\bm{x}_{s}$ denote a clean image, and $y$ is the corresponding label. $\mathcal{F}_{\phi}(\cdot)$ is the classifier with parameters $\phi$. 
The targeted attack aims to mislead the classifier to output a specific target class from an adversarial example $\bm{x'}_{s}$ corresponding to the clean image $\bm{x}_{s}$, as formulated as Eq.~\ref{eq:target_attack}.
\begin{equation}
    {\mathcal{F}_{\phi}(\bm{x'}_{s})} = y^{t}, 
    \,\,\,\,\,\,\,\,\,\,\,\,\,\,\, s.t.\,\,\|\bm{x'}_{s}-\bm{x}_{s}\|_{\infty}\leq\epsilon
\label{eq:target_attack}
\end{equation}
where $y^{t}$ represents a specific target class and it is often constrained as one of known classes, $\epsilon$ is the perturbation constraint.

\begin{figure}[tb]
  \centering
  \includegraphics[width=\linewidth]{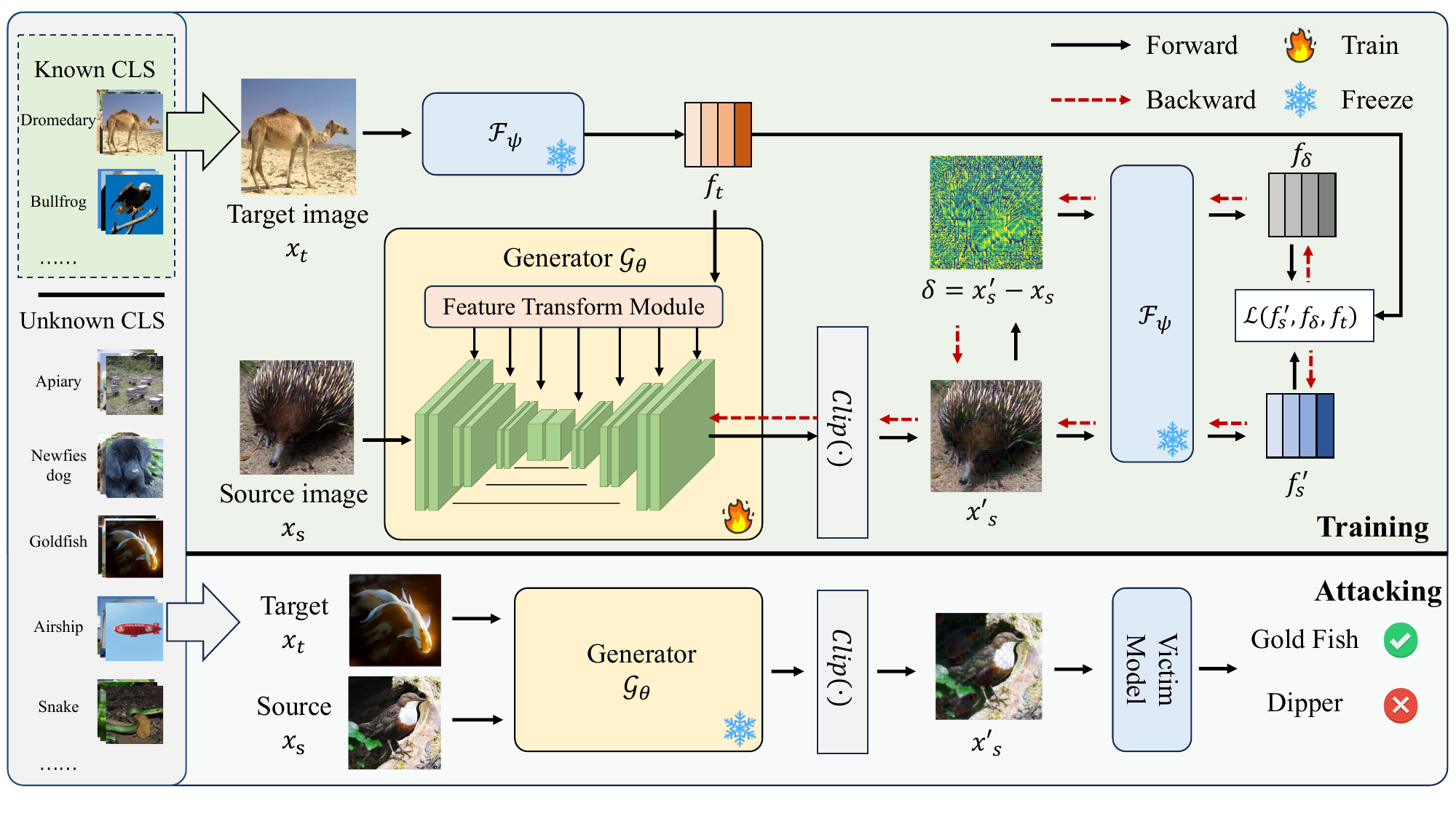}
  \caption{\textbf{The pipeline of our GAKer.} 
  We propose a generator-based method, GAKer, that can achieve attacks even when targeting classes unseen during training.
  During the training phase, we extract target features $f_t$ through a frozen $\mathcal{F}_\psi$ and inject them into the generator $\mathcal{G}_\theta$, then use $Clip(\cdot)$ to constrain $x'_s$ within the perturbation budget. $\mathcal{G}_\theta$ aims to minimize the cosine similarity between $f'_s$ and $f_t$, as well as between $f_\delta$ and $f_t$. 
  Due to our training strategy built on the feature distribution independent of the training classes, our generator can generate adversarial examples $x'_s$ for unknown classes to attack the victim model.
  }
  \label{fig:pipeline}
\end{figure}

In terms of arbitrary-target attack, it releases the constraint of the specific target class and aims to construct adversarial examples from any target regardless of whether it is from known classes or not. Given an arbitrary target image $\bm{x}_t$, the adversarial example generation is formulated as Eq.~\ref{eq:any_target_attack}:
\begin{equation}
\bm{x'}_s = \min(\bm{x}_s+\epsilon,\max(\mathcal{G}(\bm{x}_s, \bm{x}_t), \bm{x}_s-\epsilon)),
\label{eq:any_target_attack}
\end{equation}
where $\mathcal{G}_i$ denotes a trained adversarial generator. Compared with the conventional targeted attack, the arbitrary-target attack is more challenging since it requires a generator with a strong generalization capability.

\subsection{Generalized Adversarial Attacker}
\label{sec:pipeline}

In this section, we introduce the details of the proposed \textbf{Generalized Adversarial Attacker (GAKer)}. The overall pipeline is depicted in \cref{fig:pipeline}.
Given a target object, the GAKer intends to contaminate latent representations of the clean image by utilizing major components of the target. It is worth noting that 
the target object can be either a one-hot class label or an image with the visual appearance of the object. Existing methods~\cite{yang2022HGN, han2019MAN, gao2023ESMA} use the label index or the one-hot label as the condition for targeted adversarial example generation. However, the one-hot label lacks visual characteristics, which leads a trained generator to memory class features, thus limiting the generalization capability. 
To incorporate richer target information, we use images of the target object as input for generating adversarial examples.
Next, we divide classes into known classes and unknown classes $\bm{\mathcal{Y}} = \bm{\mathcal{Y}}_{\text{known}} \cup \bm{\mathcal{Y}}_{\text{unknown}}$. Then we select known classes as the training set ($t \sim \bm{\mathcal{Y}_{\text{known}}}$) and generate an adversarial example using relevant image $\bm{x}_t$ of the target. The objective function is represented as Eq.~\ref{eq:object_funtion}:
\begin{equation}
    \underset{\theta}{\min} \mathbb{E}_{(\bm{x}_s \sim \mathcal{X}_s, t \sim \mathcal{Y}_{known})}[\mathcal{L}(\bm{x}_s, \bm{x}_t)], 
\label{eq:object_funtion}
\end{equation}
where $\mathcal{L}$ is the loss function, $\mathcal{F}_{\psi}(\cdot)$ is the pretrained feature extractor, and $\mathcal{G}_\theta$ is our arbitrary-target generator (see \cref{sec:generator} for detailed architecture). 
To generalize to unknown target classes during the inference phase, 
we use cosine distance as the loss function instead of cross-entropy:
\begin{equation}
    \mathcal{D}_{cos}(f'_s, f_t) = 1 - \frac{f'_s \cdot f_t}{\left \| f'_s \right \|_2 \cdot \left \| f_t \right \|_2},
\end{equation}
where $f'_{s}=\mathcal{F}_\psi(\bm{x'}_s)$, $f_{t}=\mathcal{F}_\psi(\bm{x}_t)$.
In addition, an identical learning objective is used to constraint the feature of adversarial perturbation $\delta=\bm{x'}_s-\bm{x}_s$ and the feature of the target object:
\begin{equation}
\begin{aligned}
    \mathcal{L}(\bm{x}_s, \bm{x}_t) = & \mathcal{D}_{cos}(\mathcal{F}_\psi(\mathcal{G}_{\theta}(\bm{x}_s, \bm{x}_t)), \mathcal{F}_\psi(\bm{x}_t)) \\
    & + \alpha \mathcal{D}_{cos}(\mathcal{F}_\psi(\mathcal{G}_{\theta}(\bm{x}_s, \bm{x}_t) - \bm{x}_s), \mathcal{F}_\psi(\bm{x}_t)),
\end{aligned}
\label{eq:loss_fn}
\end{equation}
where $f_{\delta}=\mathcal{F}_\psi(\delta)$, $\alpha$ denotes the hyper-parameter (see Appendix \textcolor{red}{C} for the effect of $\alpha$). The entire training process is independent of specific target classes, enabling adaptation to unknown classes, including those from different datasets. 

\subsection{Latent Infection}

\label{sec:generator}

This section describes how to inject target features into the source image in the latent feature space. 
There are two major modules in GAKer: the Feature Extractor $\mathcal{F}_\psi$ and the Generator $\mathcal{G}_\theta$, as shown in the \cref{fig:pipeline}. The 
Feature Extractor is adapted from a pretrained model by removing the classification head. It is specialized for extracting feature vectors from input images. During the training, the weights of this module are frozen. 
We employ a UNet~\cite{gao2023ESMA} as the basic architecture of the generator.
It is designed to generate adversarial examples by using clean images with features of the target object obtained from the Feature Extractor.

\begin{wrapfigure}{t}{0.5\linewidth}
    \centering  
    \includegraphics[width=0.8\linewidth]{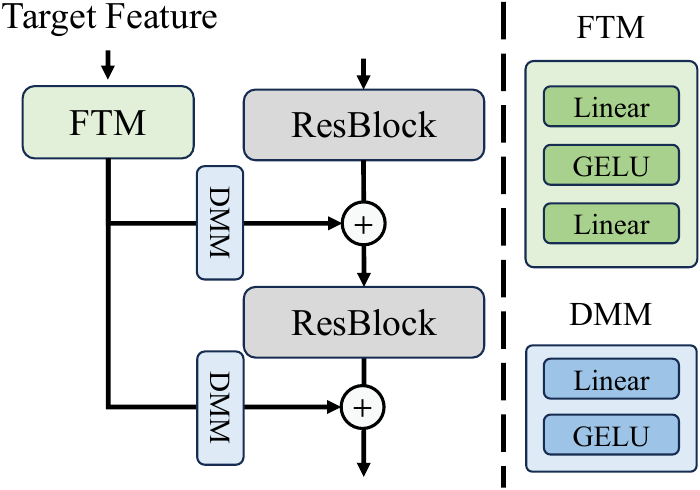}
    \caption{\textbf{Schematic diagram of feature insertion into each ResBlock of a UNet.} The features are first transformed by the Feature Transform Module (FTM), followed by dimension matching through the Dimension Matching Module (DMM) layers before being integrated into each ResBlock of the UNet.
    }
    \label{fig:generator}
\end{wrapfigure}
As depicted in \cref{fig:generator}, the latent infection involves a two-step process. First, features of the target object are processed through the Feature Transform Module (FTM), which adopts a Linear-GELU-Linear sequence to enhance their representational capacity. Second, these enhanced features are combined with the features of the clean image. Particularly, the Dimension Matching Module (DMM), which consists of a Linear-GELU layer, is used to align dimensions for two features. With the designed architecture, the generator is optimized to extract major components from the target object, and learns to replace the major components of the clean image with that of the target object.

\section{Experiment}

\subsection{Experimental Settings} \label{sec: Experimental Settings}

\noindent\textbf{Datasets.}
We train our models on the ImageNet training set~\cite{jia2009imagenet}. Correspondingly, we evaluate the performance on the ImageNet val set.

\noindent\textbf{Networks.}
We consider several models, including DenseNet-121 (Dense-121)~\cite{huang2017densely}, ResNet-50 (Res-50)~\cite{he2016deep} and VGG-19 as surrogate models. We select various black-box models, i.e., ResNet-152 (Res-152)~\cite{he2016deep}, VGG-19~\cite{simonyan2015very}, Inception-v3 (Inc-v3)~\cite{szegedy2016rethinking}, ViT~\cite{touvron2021training}, DeiT~\cite{Touvron2021DeiT} and CLIP~\cite{Radford2021learning_clip}, for testing the transferability of attacks. Additionally, we evaluate the proposed method on defense models, including Inc-$\rm v3_{adv}$, Inc-$\rm v3_{ens3}$, Inc-$\rm v3_{ens4}$, IncRes-$\rm v2_{ens}$~\cite{tramer2018ensemble}, and Large Vision-Language Models (LVLM) such as LLaVA~\cite{liu2023llava, liu2023improvedllava_llava15} and Qwen-VL~\cite{bai2023qwenvl}.

\noindent\textbf{Baselines.} For iterative attacks, we compare our method with MI~\cite{dong2018boosting_MIFGSM} and the advanced method SU~\cite{wei2023SU}, which is competitive in target settings. For single-target generative attacks, we choose TTP~\cite{naseer2021TTP} as the method for comparison with our approach. Specifically, we train multiple TTP models to accomplish multi-target attacks. For multi-target generative attacks, we choose HGN~\cite{yang2022HGN} and ESMA~\cite{gao2023ESMA}. Both of these methods, along with ours, only require training a single generator to perform attacks on multiple targets.

\noindent\textbf{Implementation details.}
In all experiments, the perturbation constraint $\epsilon$ is set to 16, the number of known classes $N$ is set to 200, the $\alpha$ is set to 0.5 and the number of samples in each known class $M$ is 325. We train the generator with an AdamW optimizer for 20 epochs. 
For the MI method, we set the decay factor $\mu$ to 1. For the SU method, we perform the combinational attack of DTMI~\cite{xie2019improving_DIFGSM, dong2018boosting_MIFGSM} and SU~\cite{wei2023SU}. For both iterative methods, we employ the logit loss and set the number of iterations to 300 steps. 
For multi-target generative attacks, such as HGN and ESMA, and our method, only one model needs to be trained to achieve attacks on multiple target classes. 
Detailed training costs and implementation specifics are provided in Appendix \textcolor{red}{A} and Appendix \textcolor{red}{B}, respectively.

\subsection{Main Results} \label{sec: main results}

\noindent\textbf{Results on Unknown Classes.} Compared with existing generator-based attacks, the best innovation of our method is the ability to attack unknown classes. We select 200 classes as known classes and the remaining 800 classes as unknown classes from ImageNet. Then we train the generator with the known classes and evaluate the targeted attack success rate on unknown classes. 
Notably, only HGN and our method can be evaluated on unknown classes. 

As shown in \cref{table:unknown_transfer}, our method significantly outperforms HGN, highlighting the superior transferability of our approach. For instance, with a substitute model of ResNet-50 and a black-box model of VGG-19, our method achieves a success rate of \textbf{41.69\%} on unknown classes, while HGN only achieves 0.05\%. This result underscores the limitation of existing methods in generating targeted adversarial examples for unknown classes, while our method demonstrates effective generalization to such classes. Separate average results on all unknown classes can be found in Appendix \textcolor{red}{D}.

\begin{table}[t]
    \centering
    \caption{\textbf{Targeted Attack Success Rates on Unknown Classes.} We report targeted attack success rates (\%) of each method and the leftmost model column denotes the substitute model (``*'' means white-box attack results). }
    \resizebox{\linewidth}{!}{
    \begin{tabular}{cc*{9}{>{\centering\arraybackslash}m{\dimexpr \linewidth/10\relax}}}
    \toprule
    \multirow{2}{*}{Model} & Attacks & Res-50 & Res-152 & VGG-19 & Dense-121 & Inc-v3 & ViT & DeiT & CLIP & Avg \\
    & Clean & 0.03 & 0.01 & 0.00 & 0.03 & 0.05 & 0.00 & 0.00 & 0.02 & 0.02\\
    \midrule
    \multirow{2}{*}{Res-50} 
        & HGN & 0.05* & 0.14 & 0.15 & 0.06 & 0.04 & 0.05 & 0.06 & 0.10 & 0.08\\
        & \cellcolor{gray!25} GAKer (Ours) & \cellcolor{gray!25} \textbf{41.69*} & \cellcolor{gray!25} \textbf{23.05} & \cellcolor{gray!25} \textbf{26.02} & \cellcolor{gray!25} \textbf{23.80} & \cellcolor{gray!25} \textbf{5.85} & \cellcolor{gray!25} \textbf{1.44} & \cellcolor{gray!25} \textbf{4.99} & \cellcolor{gray!25} \textbf{4.36} & \cellcolor{gray!25} \textbf{16.40}\\
        \midrule

        \multirow{2}{*}{VGG-19}
            & HGN & 0.08 & 0.10 & 0.08* & 0.10 & 0.04 & 0.04 & 0.04 & 0.10 & 0.07 \\
            & \cellcolor{gray!25} GAKer (Ours) & \cellcolor{gray!25} \textbf{11.05} & \cellcolor{gray!25} \textbf{5.41} & \cellcolor{gray!25} \textbf{43.25*} & \cellcolor{gray!25} \textbf{13.33} & \cellcolor{gray!25} \textbf{3.20} & \cellcolor{gray!25} \textbf{0.51} & \cellcolor{gray!25} \textbf{2.55} & \cellcolor{gray!25} \textbf{2.28} & \cellcolor{gray!25} \textbf{10.20} \\
            \midrule
            \multirow{2}{*}{Dense-121}
        & HGN & 0.06 & 0.05 & 0.09 & 0.05* & 0.00 & 0.03 & 0.01 & 0.05 & 0.04\\
        & \cellcolor{gray!25} GAKer (Ours) & \cellcolor{gray!25} \textbf{23.10} & \cellcolor{gray!25} \textbf{17.28} & \cellcolor{gray!25} \textbf{25.17} & \cellcolor{gray!25} \textbf{40.31*} & \cellcolor{gray!25} \textbf{7.23} & \cellcolor{gray!25} \textbf{2.60} & \cellcolor{gray!25} \textbf{7.56} & \cellcolor{gray!25} \textbf{4.49} & \cellcolor{gray!25} \textbf{15.97} \\

    \bottomrule
    \end{tabular}
    }
    \label{table:unknown_transfer}
\end{table}

\noindent\textbf{Results on Known Classes.} For evaluation on known classes, we compare our method with state-of-the-art iterative-based method (SU), single-target generator-based attacks (TTP), and multi-target generator-based attacks (HGN, ESMA).
All multi-target generator-based attacks (HGN, ESMA, and our method) are trained on the same 200 classes. Due to the TTP method requiring training a model for each target class, the cost of training 200 models is prohibitively high. Therefore, we randomly select 10 classes from the 200 classes and train 10 TTP models separately for each substitute model (TTP-10). We then test our method on the same 10 classes (GAKer-10). 

\Cref{table:known_transfer} shows the targeted attack success rates on known classes for each method. 
Compared with the iterative-based method SU, our method achieves higher targeted attack success rates on black-box models. For example, if the substitute model is Dense-121, 
our method performs 10.38\% better than SU on average across different models. For generator-based attacks, our GAKer also achieves a similar attack success rate to the ESMA method, outperforming the HGN method.
Notably, our method improves performance on several models by an average of 10.5\% and 5.47\% over HGN and ESMA, respectively.

\begin{table}[t]
    \centering
    \caption{\textbf{Targeted Transfer Success Rates on Known Classes}. We report targeted attack success rates (\%) of each method and the leftmost model column denotes the substitute model (``*'' means white-box attack results). }
    \resizebox{\linewidth}{!}{
    \begin{tabular}{cc*{9}{>{\centering\arraybackslash}m{\dimexpr \linewidth/10\relax}}}
    \hline
    Model & Attacks & Res-50 & Res-152 & VGG-19 & Dense-121 & Inc-v3 & ViT & DeiT & CLIP & Avg \\
    & Clean & 0.02 & 0.02 & 0.01 & 0.03 & 0.03 & 0.02 & 0.02 & 0.01 & 0.02\\
    \hline
    \multirow{6}{*}{{Res-50}}
        & MI & 99.35* & 16.55 & 4.10 & 12.10 & 0.65 & 0.60 & 0.65 & 0.15 & 16.77 \\
        & SU & \textbf{99.45*} & 83.15 & 75.60 & 81.15 & 14.95 & 8.15 & 21.80 & 6.90 & 48.89 \\
        & HGN & 87.06* & 55.68 & 52.09 & 64.59 & 24.09 & 16.36 & 29.35 & 6.71 & 41.99 \\
        & ESMA & 95.60* & 83.22 & 81.98 & \textbf{82.54} & \textbf{40.14} & \textbf{28.01} & \textbf{55.45} & \textbf{24.08} & \textbf{61.37} \\
        & \cellcolor{gray!25} GAKer (Ours) & \cellcolor{gray!25} 96.61* & \cellcolor{gray!25} \textbf{83.36} & \cellcolor{gray!25} \textbf{82.20} & \cellcolor{gray!25} 81.95 & \cellcolor{gray!25} 34.27 & \cellcolor{gray!25} 20.84 & \cellcolor{gray!25} 50.31 & \cellcolor{gray!25} 20.13 & \cellcolor{gray!25} 58.71 \\
        \cmidrule{2-11}
        & TTP-10 & 97.80* & 77.80 & 73.00 & 79.00 & 44.00 & \textbf{33.10} & 44.00 & 19.40 & 58.51 \\
        & \cellcolor{gray!25} GAKer-10 (Ours) & \cellcolor{gray!25} \textbf{98.10*} & \cellcolor{gray!25} \textbf{88.20} & \cellcolor{gray!25} \textbf{90.60} & \cellcolor{gray!25} \textbf{86.00} & \cellcolor{gray!25} \textbf{45.20} & \cellcolor{gray!25} 28.70 & \cellcolor{gray!25} \textbf{62.00}  & \cellcolor{gray!25} \textbf{24.50} & \cellcolor{gray!25} \textbf{65.41}  \\
        \hline
    \multirow{6}{*}{{VGG-19}}
        & MI & 3.10 & 1.40 & 96.85* & 3.60 & 0.30 & 0.20 & 0.55 & 0.35 & 13.29 \\
        & SU & 26.35 & 14.29 & 96.90* & 26.05 & 4.10 & 2.20 & 5.89 & 4.60 & 22.55 \\
        & HGN & 28.50 & 16.65 & 90.15* & 33.05 & 6.95 & \textbf{6.00} & 13.50 & 2.95 & 24.72 \\
        & ESMA & 27.30 & 17.85 & 95.39* & 35.80 & 4.85 & 5.35 & \textbf{16.70} & \textbf{6.12} & 26.17 \\
        & \cellcolor{gray!25} GAKer (Ours) & \cellcolor{gray!25} \textbf{30.00} & \cellcolor{gray!25} \textbf{18.02} & \cellcolor{gray!25} \textbf{97.61*} & \cellcolor{gray!25} \textbf{33.38} & \cellcolor{gray!25} \textbf{10.76} & \cellcolor{gray!25} 3.36 & \cellcolor{gray!25} 11.65 & \cellcolor{gray!25} 4.75 & \cellcolor{gray!25} \textbf{26.22} \\
        \cmidrule{2-11}
        & TTP-10 & \textbf{50.10} & \textbf{37.50} & \textbf{98.20*} & \textbf{48.80} & \textbf{17.20} & \textbf{13.40} & \textbf{20.00} & \textbf{10.60} & \textbf{36.98} \\
        & \cellcolor{gray!25} GAKer-10 (Ours) & \cellcolor{gray!25} 
        32.00 & \cellcolor{gray!25} 16.50 & \cellcolor{gray!25} 93.80* & \cellcolor{gray!25} 32.70 & \cellcolor{gray!25} 12.00 & \cellcolor{gray!25} 3.05 & \cellcolor{gray!25} 10.05 & \cellcolor{gray!25} 4.00 & \cellcolor{gray!25} 25.51
        \\
        \hline
    \multirow{6}{*}{Dense-121}
        & MI & 10.25 & 5.85 & 5.15 & \textbf{99.55*} & 1.00 & 0.65 & 1.20 & 0.50 & 15.52 \\
        & SU & 65.20 & 49.70 & 59.90 & \textbf{99.55*} & 14.90 & 7.65 & 19.35 & 6.00 & 40.28 \\
        & HGN & 53.48 & 40.15 & 50.67 & 90.11* & 23.32 & 13.92 & 22.35 & 5.00 & 37.38 \\
        & ESMA & 43.01 & 34.74 & 42.61 & 81.52* & 11.74 & 9.89 & 23.47 & 6.24 & 31.65 \\
        & \cellcolor{gray!25} GAKer (Ours) & \cellcolor{gray!25} \textbf{70.26} & \cellcolor{gray!25} \textbf{61.12} & \cellcolor{gray!25} \textbf{69.83} & \cellcolor{gray!25} 90.91* & \cellcolor{gray!25} \textbf{30.81} & \cellcolor{gray!25} \textbf{22.15} & \cellcolor{gray!25} \textbf{43.81} & \cellcolor{gray!25} \textbf{16.41} & \cellcolor{gray!25} \textbf{50.66} \\
        \cmidrule{2-11}
        & TTP-10 & \textbf{86.70} & \textbf{76.50} & 82.80 & \textbf{97.90*} & \textbf{49.20} & \textbf{44.10} & 57.00 & \textbf{24.60} & \textbf{64.85} \\
        & \cellcolor{gray!25} GAKer-10 (Ours) & \cellcolor{gray!25} 74.40 & \cellcolor{gray!25} 70.20 & \cellcolor{gray!25} \textbf{82.89} & \cellcolor{gray!25} 93.20* & \cellcolor{gray!25} 42.80 & \cellcolor{gray!25} 28.99 & \cellcolor{gray!25} \textbf{60.00} & \cellcolor{gray!25} 23.10 & \cellcolor{gray!25} 59.45 \\
    \bottomrule
    \end{tabular}
    }
    \label{table:known_transfer}
\end{table}

\subsection{Results on Other Models} \label{Generalization Study}

To further evaluate the generalization ability of our method, we also test our method on defense models and Large Vision-Language Models (LVLM).

\noindent\textbf{Results on Defense Models.} In addition to attacking normally trained models, we evaluate the attack performance of various methods on adversarially trained models when using ResNet50 as the white-box model. 
As shown in \cref{table:defense_models}, our method surpasses the HGN method in attacking targets in unknown and known classes.
For example, with the defense model as Inc-v$3_{adv}$, we outperform HGN by \textbf{6.46\%} in terms of target success rate on the known classes. On the unknown classes, HGN cannot achieve an attack at all, similar to the performance of the clean samples, whereas our GAKer still exhibits an attack success rate of \textbf{5.95\%}. This result indicates that our generator has discovered more common model vulnerabilities, regardless of whether the model has been specifically trained for defense.

\begin{table}[t]
    \centering
    \caption{\textbf{Targeted Transfer Success Rates on Defense Models}. Targeted success rates (\%) for each attack method using Res-50 as the substitute model (known classes / unknown classes).}
    \begin{tabular}{c*{4}{>{\centering\arraybackslash}p{\dimexpr 0.7\linewidth/4\relax}}}
    \toprule
     Attacks & Inc-v$3_{adv}$ & Inc-v$3_{ens3}$ & Inc-v$3_{ens4}$ & IncRes-v$2_{ens}$ \\
    \midrule
    Clean &  0.03/0.04 & 0.04/0.03 & 0.02/0.01 & 0.03/0.03 \\
    HGN   & 29.22/0.06 & 26.24/0.06 & \textbf{24.88}/0.07 & 21.69/0.04 \\
    \rowcolor{gray!25} GAKer (Ours) & \textbf{35.68}/\textbf{5.95} & \textbf{29.60}/\textbf{4.10} & 23.66/\textbf{3.38} & \textbf{25.25}/\textbf{3.21} \\
    \bottomrule
    \end{tabular}
    \label{table:defense_models}
\end{table}

\noindent\textbf{Results on Large Vision-Language Models.}
The growing importance of Large Vision-Language Models (LVLM), including LLaVA~\cite{liu2023llava, liu2023improvedllava_llava15} and Qwen-VL~\cite{bai2023qwenvl}, has been noted in recent times. Our research examines how well our method can be adapted for use with these LVLMs.
Specifically, we test multiple templates on LVLMs to reduce the impact of prompt bias:
\begin{itemize}
    \item Is there any \underbar{(origin class / target class)} in this image? Please begin answer with `Yes,' or `No,'
    \item Does this image contain any \underbar{(origin class / target class)}? Please begin answer with `Yes,' or `No,'
    \item In this image, is there a \underbar{(origin class / target class)} present? Please begin answer with `Yes,' or `No,'
\end{itemize}

\begin{table}[t]
    \centering
    \caption{\textbf{Attack Success Rates for Unknown Classes on LVLM}. We report attack success rates (\%) of each method and the topmost row denotes the substitute model. (untargeted attack success rates / targeted attack success rates)}
    \resizebox{\linewidth}{!}{
    \begin{tabular}{c*{6}{>{\centering\arraybackslash}p{\dimexpr 1\linewidth/6\relax}}}
    \toprule
     {Substitute Models} & \multicolumn{2}{c}{Res-50} & \multicolumn{2}{c}{Dense-121} & \multicolumn{2}{c}{VGG-19} \\
    \cmidrule(lr){2-3}  \cmidrule(lr){4-5}  \cmidrule(lr){6-7}
     Black-box & Qwen-VL & LLaVA & Qwen-VL & LLaVA & Qwen-VL & LLaVA \\
    \midrule
     Clean & 5.77/8.15 & 9.03/7.02 & 5.77/8.15 & 9.03/7.02 & 5.77/8.15 & 9.03/7.02 \\
    \midrule
     SU & 20.16/35.36 & 18.77/42.73 & 19.59/28.76 & 17.83/32.73 & 20.40/33.35 & 18.13/\textbf{46.80} \\
     HGN & 24.98/12.50 & 20.76/13.85	& 22.12/12.27 &	20.76/13.85 & 23.65/12.79 & 20.23/14.34 \\
    \rowcolor{gray!25} \cellcolor{white}  GAKer (Ours) & \textbf{33.82}/\textbf{52.60} & \textbf{23.41}/\textbf{56.45} & \textbf{36.71}/\textbf{54.43} & \textbf{27.54}/\textbf{60.41} & \textbf{39.86}/\textbf{36.31} & \textbf{24.50}/41.61 \\

    \bottomrule
    \end{tabular}
    }
    \label{table:unknown_transfer_on_LVLM}
\end{table}

We then assessed the effectiveness of our approach by calculating both the untargeted and targeted attack success rates against LVLMs.  
We define the untargeted attack success rate as the percentage of adversarial examples that mislead the model into failing to recognize the original class. The targeted attack success rate is defined as the percentage of adversarial examples that are misclassified into the target class. As shown in \cref{table:unknown_transfer_on_LVLM} and \cref{table:known_transfer_on_LVLM}, our method achieves higher targeted attack success rates on LVLM than HGN. For example, when the substitute model is Res-50 and the target is unknown classes, our method achieves \textbf{52.60\%} and \textbf{56.45\%} targeted attack success rates on LLaVA and Qwen-VL, respectively, while HGN only achieves 12.50\% and 13.85\%. 
This result demonstrates that even when using a ``small'' substitute model like Res-50, our method can successfully attack ``large'' models such as Qwen-VL and LLaVA. We also show some cases on GPT-4V\cite{GPT4V}, which can be found in Appendix \textcolor{red}{E}.

\begin{table}[t]
    \centering
    \caption{\textbf{Attack Success Rates for Known Classes on LVLM}. We report attack success rates (\%) of each method and the topmost row denotes the substitute model. (untargeted attack success rates / targeted attack success rates)}
    \resizebox{\linewidth}{!}{
    \begin{tabular}{c*{6}{>{\centering\arraybackslash}p{\dimexpr 1\linewidth/6\relax}}}
    \toprule
    
     {Substitute Models} & \multicolumn{2}{c}{Res-50} & \multicolumn{2}{c}{Dense-121} & \multicolumn{2}{c}{VGG-19} \\
    \cmidrule(lr){2-3}  \cmidrule(lr){4-5}  \cmidrule(lr){6-7}
     Black-box & Qwen-VL & LLaVA & Qwen-VL & LLaVA & Qwen-VL & LLaVA \\
    \midrule
     Clean & 5.15/10.33 & 8.43/7.42 & 5.15/10.33 & 8.43/7.42 & 5.15/10.33 & 8.43/7.42 \\
    \midrule
     SU & 20.10/42.28 & 18.78/46.02 & 19.70/35.67 & 17.13/37.28 & 20.53/38.77 & 18.38/\textbf{48.88} \\
     HGN & 22.95/36.57 & 18.67/34.35 & 21.70/31.13 & 18.22/29.78 & 22.80/28.87 & 16.15/30.95 \\
     ESMA & \textbf{50.07}/70.53 & \textbf{33.67}/66.05 & 40.60/41.73 & \textbf{28.97}/31.12 & 38.68/\textbf{54.02} & 23.48/45.75 \\
     \rowcolor{gray!25} \cellcolor{white}  GAKer (Ours)& 35.20/\textbf{73.62} & 24.32/\textbf{73.37} & \textbf{47.60}/\textbf{63.10} & 27.30/\textbf{72.00} & \textbf{45.00}/42.50 & \textbf{26.95}/44.90 \\

    \bottomrule
    \end{tabular}
    }
    \label{table:known_transfer_on_LVLM}
\end{table}

\subsection{Ablation Study} \label{Sec: Ablation Study}
This section discusses how different parameter selections in training dataset construction impact the generator's attack capability.

\begin{wrapfigure}{t}{0.5\linewidth}
    \centering
    \includegraphics[width=0.9\linewidth]{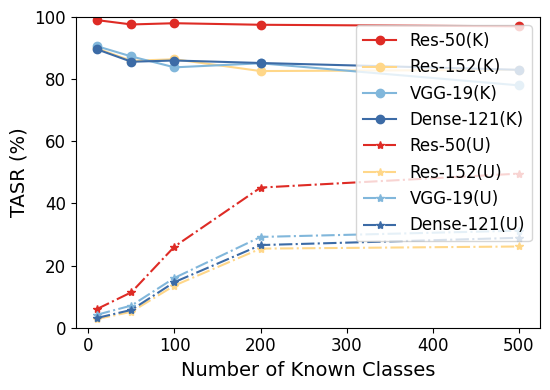}
    \caption{Comparison of targeted attack success rates across a range of known classes. The Res-50 serves as the substitute model, while the performance of black-box models, including Res-152, VGG-19, and Dense-121, is evaluated for both known (K) and unknown (U) classes.
    }
    \label{fig:classes}
\end{wrapfigure}

\noindent\textbf{Numbers of Known Classes.} \Cref{fig:classes} demonstrates 
the impact of the number N of known classes on the generator's performance. 
Specifically, to mitigate
the impact of adding new classes as the number of known classes increases, we evaluate targeted attack success rate on a common set of 10 known classes and 500 common unknown classes.
When N is 10 or 50, the attack success rate on unknown classes is low due to limited training data. With N increased to 500, the white-box success rate reaches 49.45\% on 500 unknown classes. Despite higher training costs, performance does not significantly improve with more known classes. Therefore, N is set to 200 to balance performance and training cost.

When N is 10 or 50, the attack success rate on unknown classes is low due to limited training data. With N increased to 500, the white-box success rate reaches 49.45\% on 500 unknown classes. Despite higher training costs, performance does not significantly improve with more known classes. Therefore, N is set to 200 to balance performance and cost.

\noindent\textbf{Strategies for Choosing Known Classes.} Previous work~\cite{yang2022HGN} on multi-target generators has highlighted the importance of not only determining the number of target classes but also selecting which target classes to attack. 
When there are significant differences between the selected target classes, the generator can achieve a better targeted attack success rate.

We introduce a similarity greedy algorithm to select a set of classes with the largest feature differences. To simplify, we represent each target class using the average of its image features and measure the similarity using cosine similarity. 
Specifically, the algorithm starts by randomly selecting an initial feature vector as the first class and adds it to the selected group. It then iteratively selects a vector from the remaining pool that has the lowest average cosine similarity to the vectors in the selected group.
We add this selected class to the selected group 
and repeat the process until achieve the specified number of target classes.
\begin{wrapfigure}{r}{0.5\linewidth}
    
    \centering
    \includegraphics[width=\linewidth]{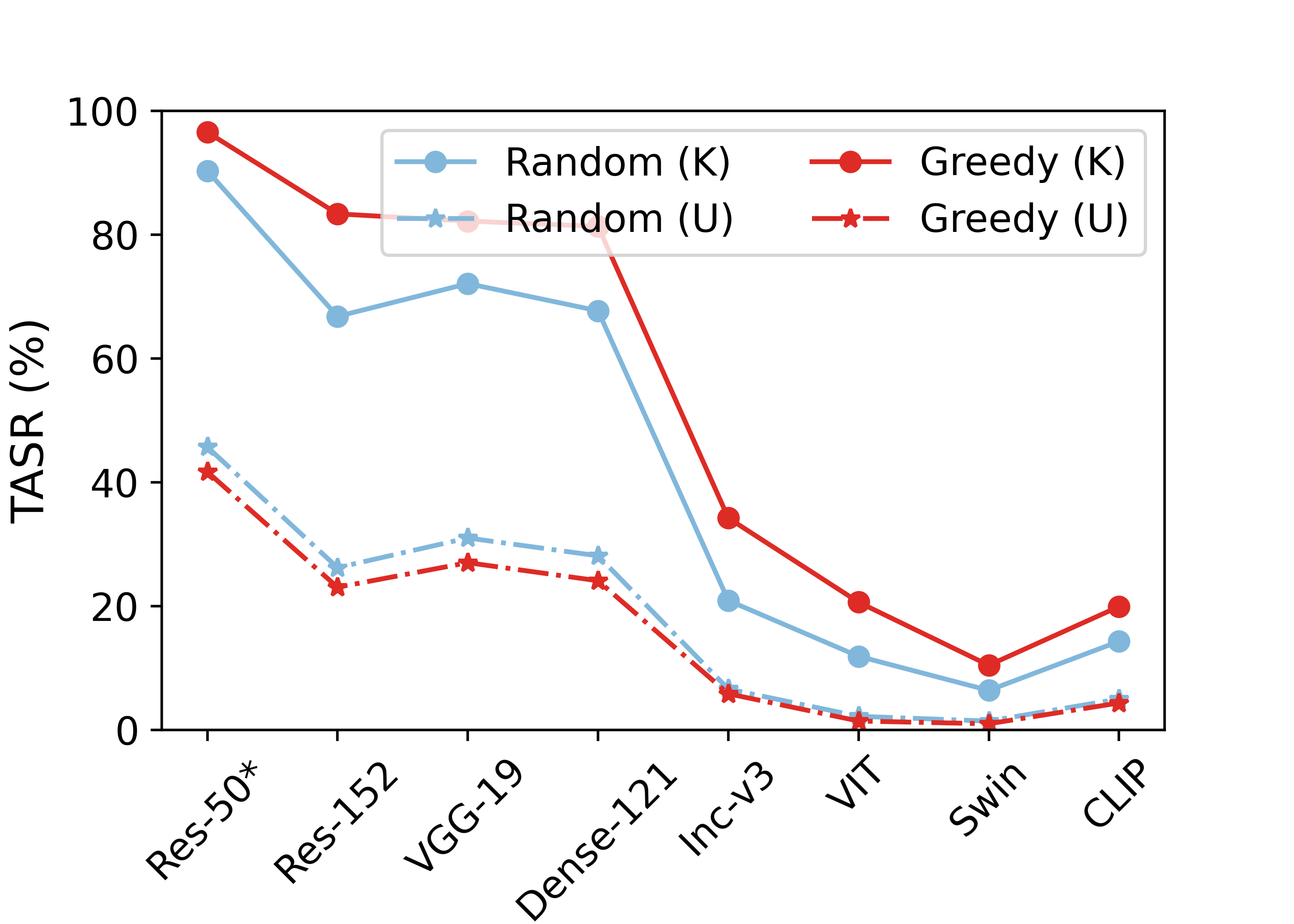}
    \caption{Comparison of targeted attack transfer success rates under different known classes selection strategies.
    }
    \label{fig:selection}

\end{wrapfigure} 

We conduct multiple experiments to eliminate the randomness introduced by the initial selection. To validate our selection method's effectiveness, we compare it with randomly selecting strategy. 
\Cref{fig:selection} illustrates that our method significantly outperforms random selection on known classes. Compared with the random strategy, we achieve a success rate increase of \textbf{16.52\%} on black-box model Res-152 when the substitute model is Res-50. This demonstrates the necessity of selecting known classes and the effectiveness of our selecting method.

\begin{figure}[t]
    \centering
    \includegraphics[width=\linewidth]{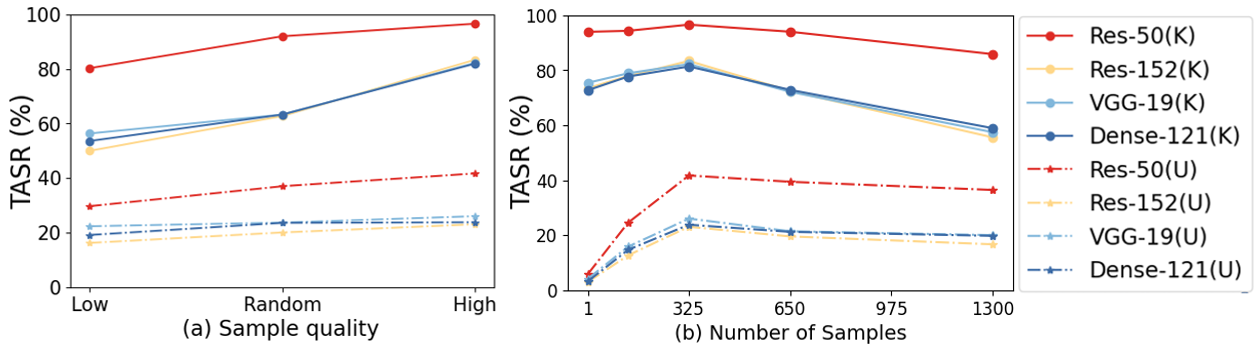}
    \caption{
    Comparisons of targeted attack success rates at sample quality and different numbers of targeted images per known class. 
    }
    \label{fig:quality_samplenum}
\end{figure}

\noindent\textbf{Numbers of Sample in Each Known Class. }
We observe that noisy images, such as those with occlusion or blurring, can decrease performance. 
To validate this, we employ three selection methods to form groups, each comprising 325 images:
 low quality (largest classification loss), high quality (lowest loss), and random. \Cref{fig:quality_samplenum}(a) demonstrates that image quality significantly impacts the attack success rate. Consequently, we prioritize training images by their classification loss, favoring those with the lowest losses. Specifically, we experiment with different numbers of images ($M \in \{1, 130, 325, 650, 1300\}$) for each known class, as illustrated in \Cref{fig:quality_samplenum}(b). 
When the number of images (M) is less than 325, the performance is lower, likely because the generator could not capture the full breadth of class characteristics. Conversely, when M exceeds 325, the attack success rate slightly decreased, potentially due to the inclusion of more noisy images. Thus, we ultimately select 325 as the optimal number of samples for each known class. This choice yields a maximum attack success rate of 41.69\% for the unknown class in the ResNet-50 model. This finding underscores the equal importance of both the quantity and quality of training samples.

\section{Conclusion}
Generator-based targeted attacks are able to mislead DNNs to any target they have been trained on, showing their dangers. For the first time, we find that the attack also extends to untrained unknown classes, and the extent of its potential harm is revealed. We propose the Generalized Adversarial attacker, which injects target feature into adversarial examples to attack unknown classes. 
Through comprehensive experiments across standard, defense, and large vision-language models, we demonstrate that our method can effectively attack unknown and known classes across models. We hope our work will draw attention to the potential dangers of generator-based targeted attacks and inspire future research in this area.

\section{Societal Impacts \& Limitation.}

\,\,\,\,\,\,\,\,\textbf{Societal Impacts. }
Previous research on generator-based target attacks requires the attacker to know the target class. Our proposed algorithm allows successful targeted attacks without this information, highlighting the risk of relying solely on dataset and model confidentiality for security. Moreover, our method’s success on unknown classes reveals inherent model vulnerabilities, offering new insights for advancing security.

\textbf{Limitation. }
While our method validates the possibility of attacking unknown classes using generator-based methods, the gap in target attack success rates between known and unknown classes still exists. In the future, we will focus on analyzing the reasons for this difference.

\section*{Acknowledgements}
This study is supported by grants from the National Natural Science Foundation of China (Grant No. 62122018, No. 62020106008, No. U22A2097, No. U23A20315), Kuaishou, and SongShan Laboratory YYJC012022019. It is also supported by the Postdoctoral Fellowship Program of CPSF under Grant Number GZB20240114.

%
%
\bibliographystyle{splncs04}
\bibliography{main}

\appendix

\newpage
\begin{center}
  \vspace*{2\baselineskip}
  \Large\bf{Any Target Can be Offense: Adversarial Example Generation via Generalized Latent Infection \\
(Supplementary Material)}
\end{center}
\renewcommand\thesection{\Alph{section}}

\section{Implement Details}

During training, we consider samples with small classification loss to be excellent target samples with more prominent features. We continue to use this estimation during testing. 

\noindent\textbf{Test on Known Classes.}
When evaluating the target attack success rate on known classes, since the attacker entirely constructs the training set of known classes, we select the target sample with the smallest classification loss in the train set as the target for that known class. 

\noindent\textbf{Test on Unknown Classes.}
When testing on unknown classes, the attacker obviously cannot access the complete target dataset to select the best sample but can still choose some target samples that are relatively clear and unobstructed. To simulate this process, we select 10 images with relatively small classification losses for each target class from the validation set of ImageNet as targets. During testing, the target samples of unknown classes are \textbf{randomly} selected from these 10 images.

\section{Efficiency Analysis}
In this section, we analyze the efficiency of different methods. To simulate the scenario of unknown classes, we divide the ImageNet-1k dataset into 5 parts, with each part appearing sequentially. Unlike other generator-based methods that require retraining for each new part, our GAKer can attack any target class with just one training process. As shown in \cref{tab:training_time}, our method only requires 1\% of the training time compared to TTP, highlighting the remarkable efficiency of our GAKer.
\begin{table}[h] 
\centering
    \caption{Training time comparison assuming ImageNet is divided into 5 parts, with 200 new target classes added each time. ``5*20*3.7h'' means retraining 5 times, each with 20 epochs, taking 3.7 hours per epoch ( in one NVIDIA GTX 4090 GPU).}
    \begin{tabular}{c|cccc}
        \toprule
    Method & TTP & HGN & ESMA  & Ours \\
        \midrule
    Time &  1k*1*13.1h(100\%) &5*20*3.7h(3\%)& 5*20*6.8h(5\%)  & 1*20*6.6h(1\%)\\
        \bottomrule
    \end{tabular}    
    \label{tab:training_time}

\end{table}

\section{Parameter Sensitivity Analysis}

\begin{figure}[h]
  \centering
  \includegraphics[width=0.62\linewidth]{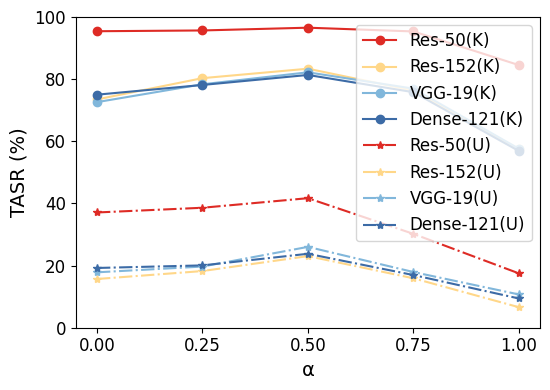}
    \caption{Effect of $\alpha$. The Res-50 serves as the substitute model, while the performance of black-box models, including Res-152, VGG-19, and Dense-121, is evaluated for both known (K) and unknown (U) classes.}
  \label{fig:alpha}
\end{figure}

To explore the impact of the parameter $\alpha$, we conduct several experiments. We train the model with $\alpha$ set to 0, 0.25, 0.5, 0.75, and 1, respectively. Then, we validate the effect of different $\alpha$ settings on the attack success rate.
In this experiment, we use ResNet-50 (Res-50)~\cite{he2016deep} as the substitute model and test the target attack success rates (TASR) on known and unknown classes on ResNet-50 (Res-50)~\cite{he2016deep}, ResNet-152 (Res-152)~\cite{he2016deep}, VGG-19~\cite{simonyan2015very}, DenseNet-121 (Dense-121)~\cite{huang2017densely}. The experimental results are shown in \cref{fig:alpha}. Finally, we select $\alpha$ as 0.5, which yields the best experimental results.

\section{Further Analysis on Unknown Classes}
\begin{figure}[h]
  \centering
  \includegraphics[width=0.5\linewidth]{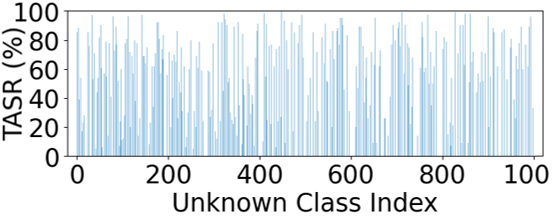}
    \caption{The TASR for each unknown class. The substitute model is Res-50.}
  \label{fig:unknown_class_index}
\end{figure}

\begin{figure}[h]
  \centering
  \includegraphics[width=0.5\linewidth]{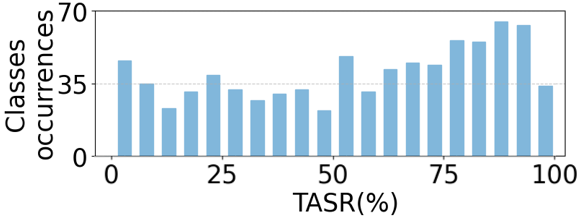}
    \caption{The histogram of unknown class distributions by TASR. The substitute model is Res-50.}
  \label{fig:class_occur}
\end{figure}

To further analyze the performance of our GAKer on unknown classes, we visualize the targeted attack success rate (TASR) for all 800 unknown classes in \cref{fig:unknown_class_index}. Meanwhile, we present the histogram of TASR in \cref{fig:class_occur}. The statistics show that there is a large variation in TASR among different unknown classes. The TASR for most of the unknown classes exceeds 50\%, but some classes are hardly attacked, with a TASR lower than 50\%. This phenomenon indicates that different classes have varying levels of difficulty.

\section{Results on GPT-4V}

We conduct a series of tests on the GPT-4V~\cite{GPT4V} to showcase the effectiveness of our method when dealing with Large Vision-Language Models. All experiments on GPT-4V are conducted on 13 March 2024. 

\begin{figure}[h]
  \centering
  \includegraphics[width=\linewidth]{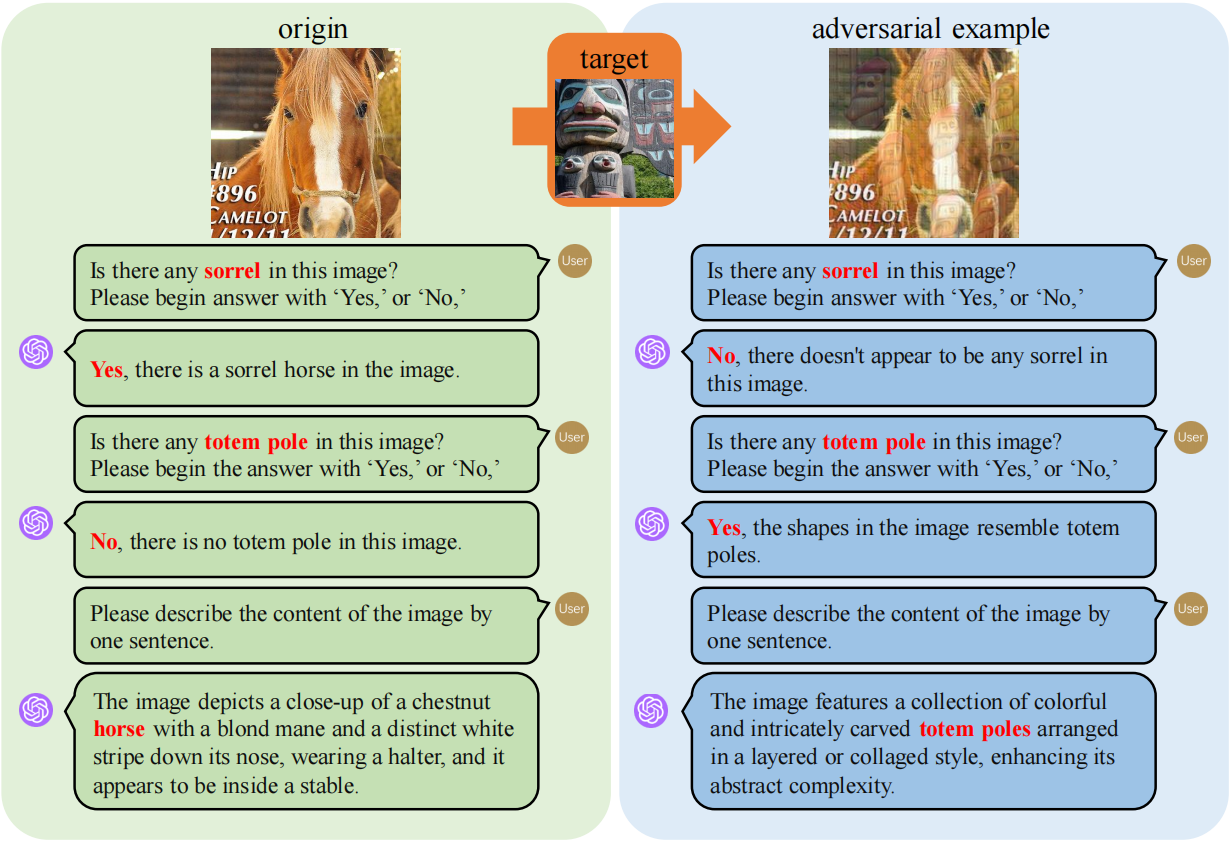}
  \caption{Q\&A from \textbf{"sorrel"} to  \textbf{"totem pole"}
}
  \label{fig:gpttalk1}
\end{figure}

\begin{figure}[t]
  \centering
  \includegraphics[width=\linewidth]{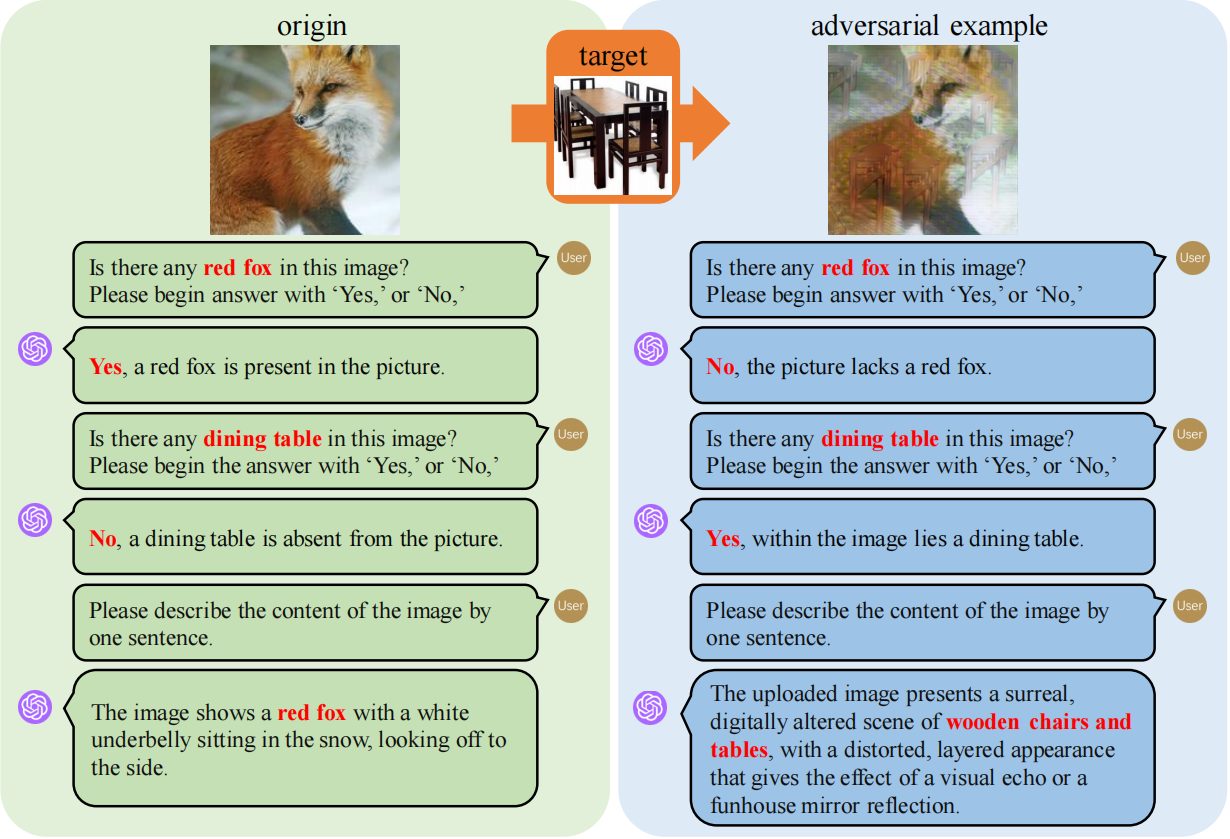}
  \caption{Q\&A from \textbf{"red fox"} to  \textbf{"dining table, board"}
}
  \label{fig:gpttalk2}
\end{figure}

\begin{figure}[t]
  \centering
  \includegraphics[width=\linewidth]{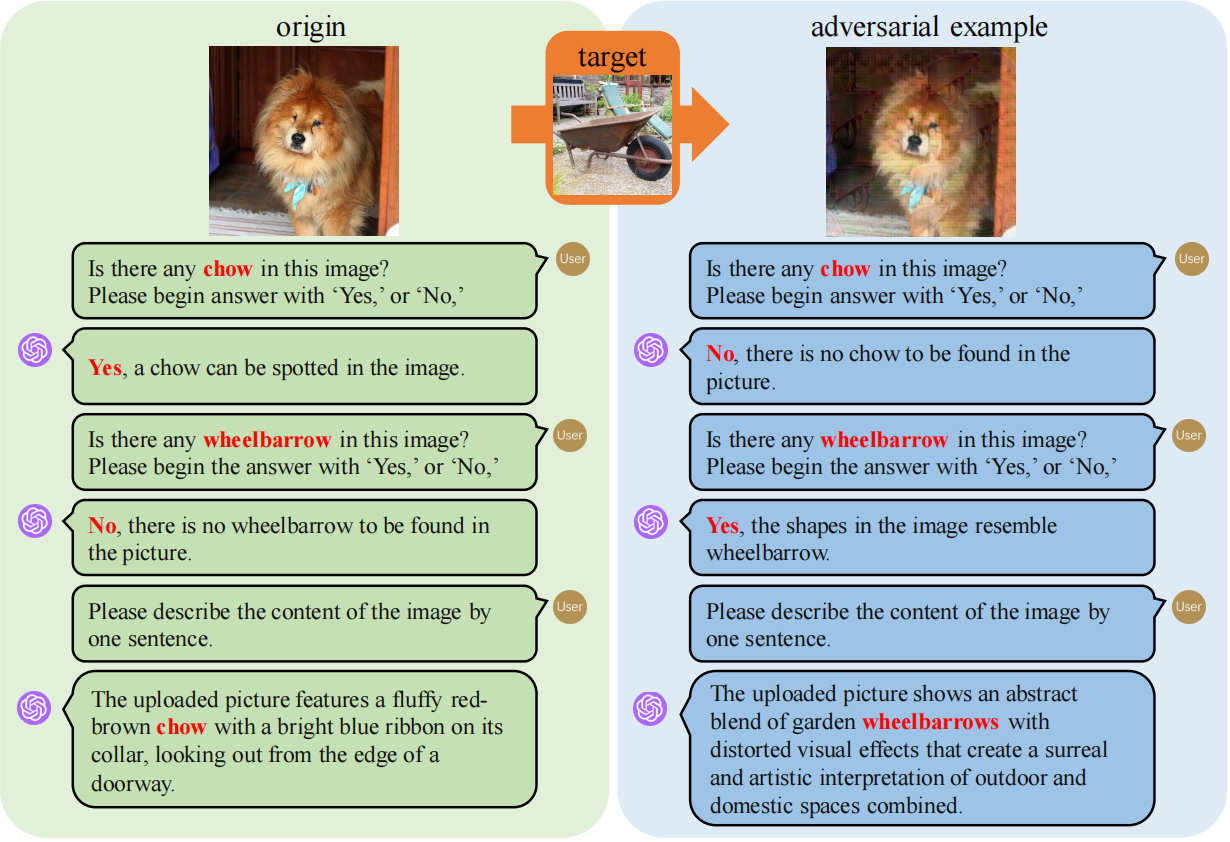}
  \caption{Q\&A from \textbf{"chow"} to  \textbf{"wheelbarrow" }
}
  \label{fig:gpttalk3}
\end{figure}

\begin{figure}[t]
  \centering
  \includegraphics[width=\linewidth]{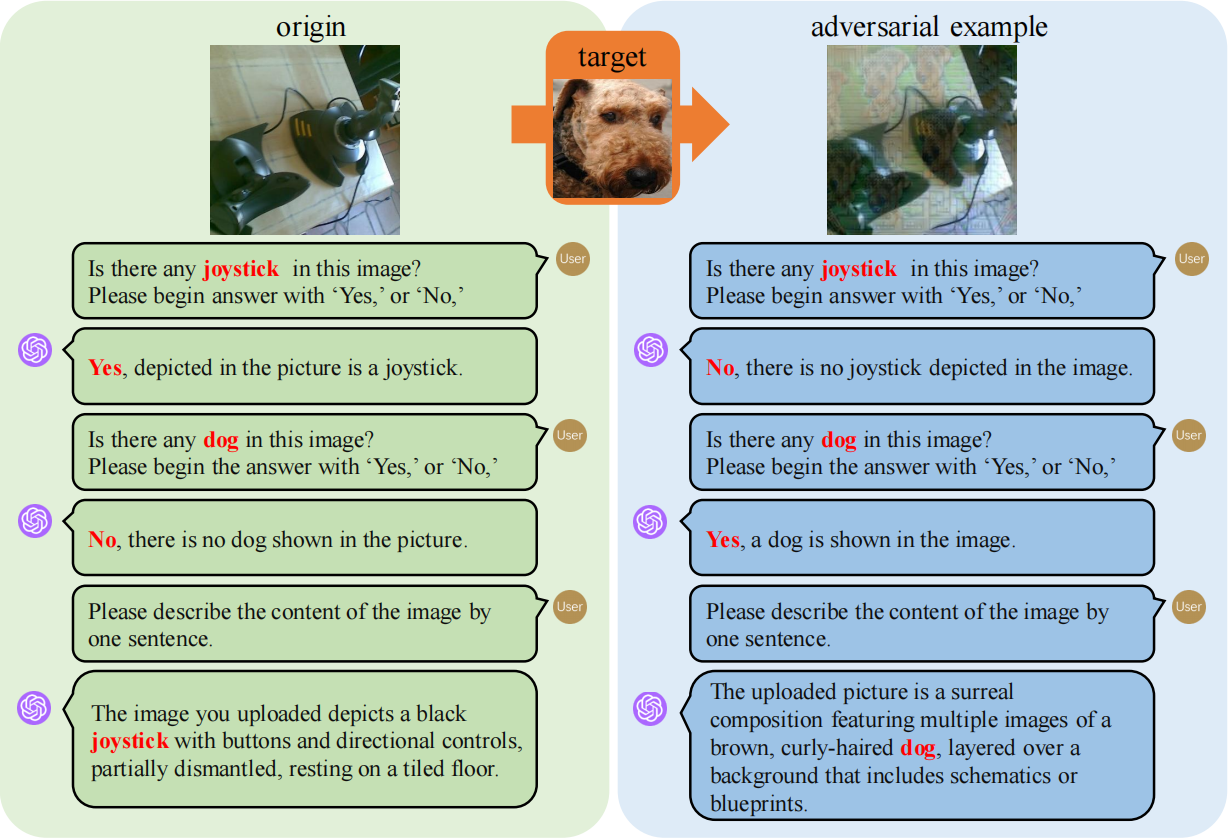}
  \caption{Q\&A from \textbf{"joystick"} to  \textbf{"dog"}}
  \label{fig:gpttalk4}
\end{figure}

As shown in \cref{fig:gpttalk1,fig:gpttalk2,fig:gpttalk3,fig:gpttalk4}, when we ask GPT-4V whether the adversarial example contains the original class or the target class, GPT-4V will deny the former and affirm the latter. Furthermore, when we ask GPT-4V to describe the content of the image by one sentence, the description corresponds with the target rather than the original image.

\begin{figure}[t]
  \centering
  \includegraphics[width=\linewidth]{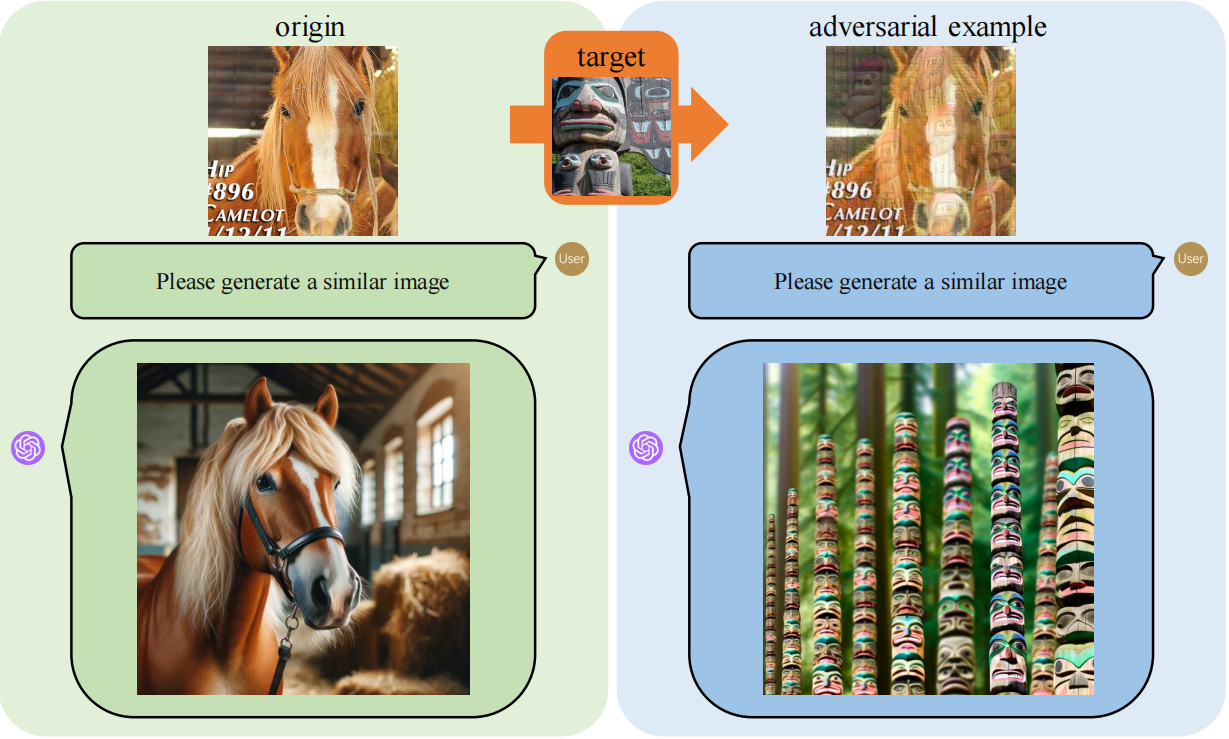}
  \caption{
  From \textbf{"sorrel"} to  \textbf{"totem pole"} generated by GPT-4V
}
  \label{fig:gpt1}
\end{figure}

\begin{figure}[t]
  \centering
  \includegraphics[width=\linewidth]{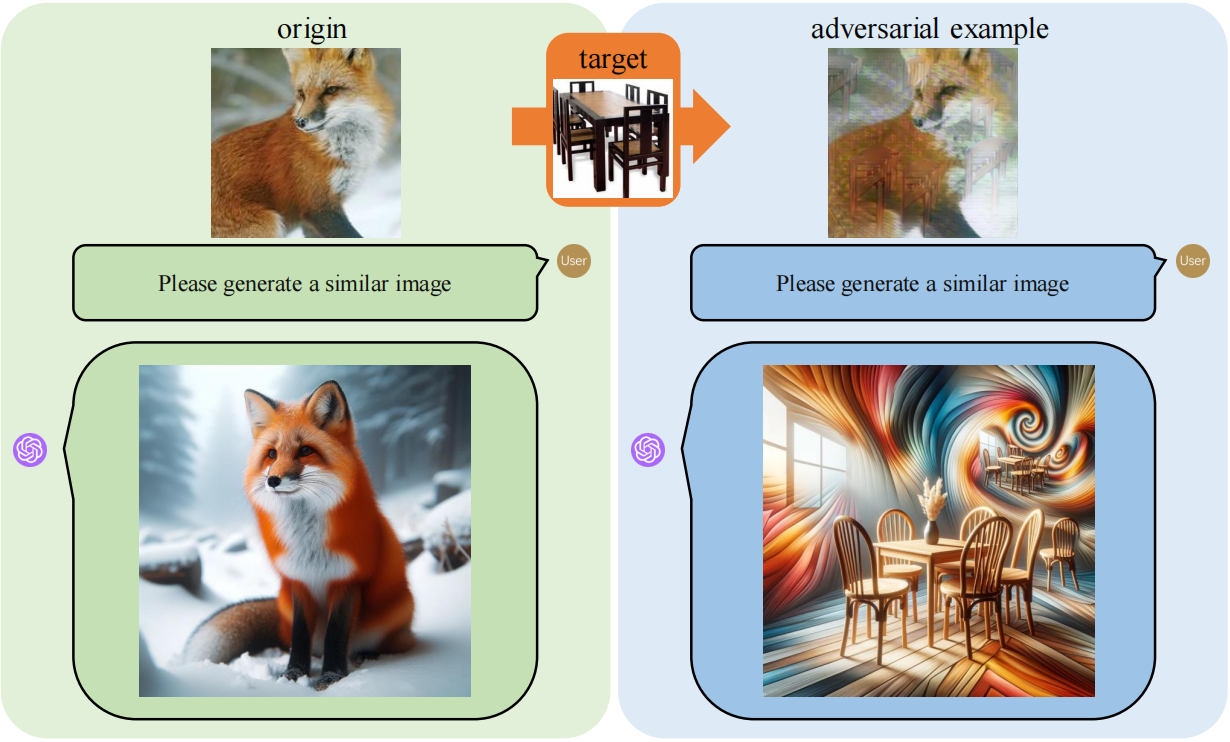}
  \caption{From \textbf{"red fox"} to  \textbf{"dining table, board"} generated by GPT-4V
}
  \label{fig:gpt2}
\end{figure}

\begin{figure}[t]
  \centering
  \includegraphics[width=\linewidth]{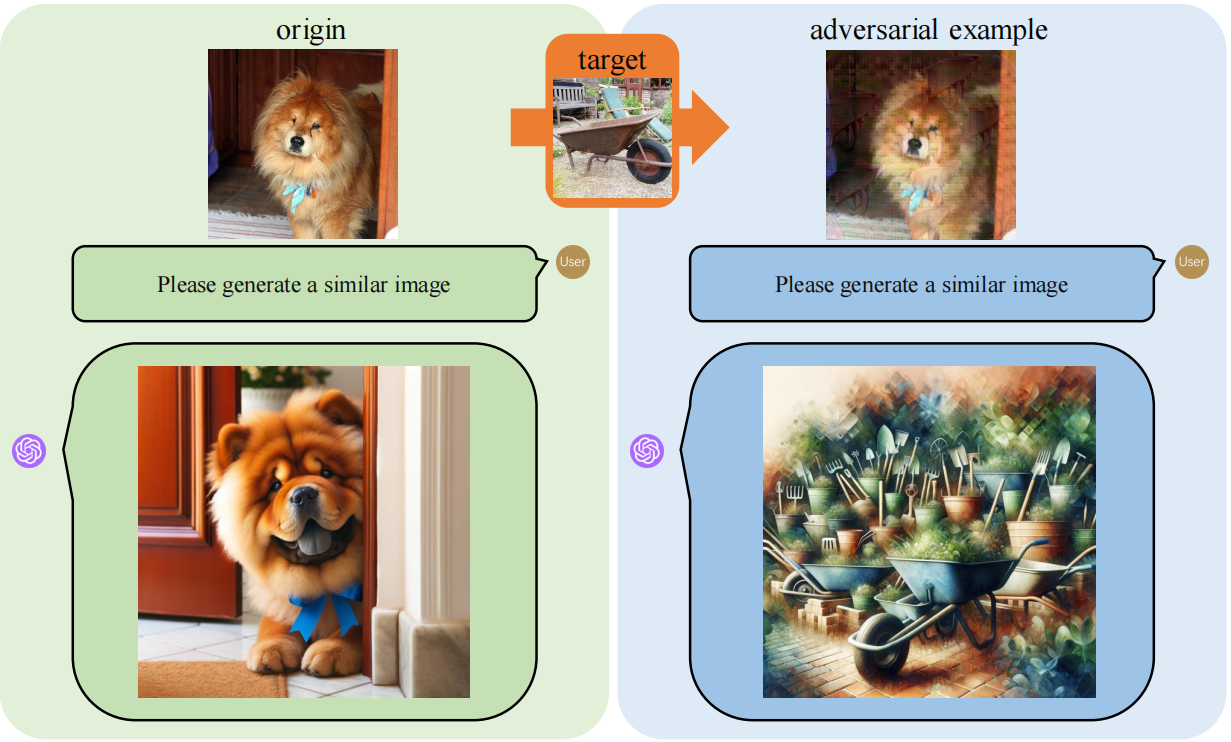}
  \caption{From \textbf{"chow"} to  \textbf{"wheelbarrow" } generated by GPT-4V
}
  \label{fig:gpt3}
\end{figure}

\begin{figure}[t]
  \centering
  \includegraphics[width=\linewidth]{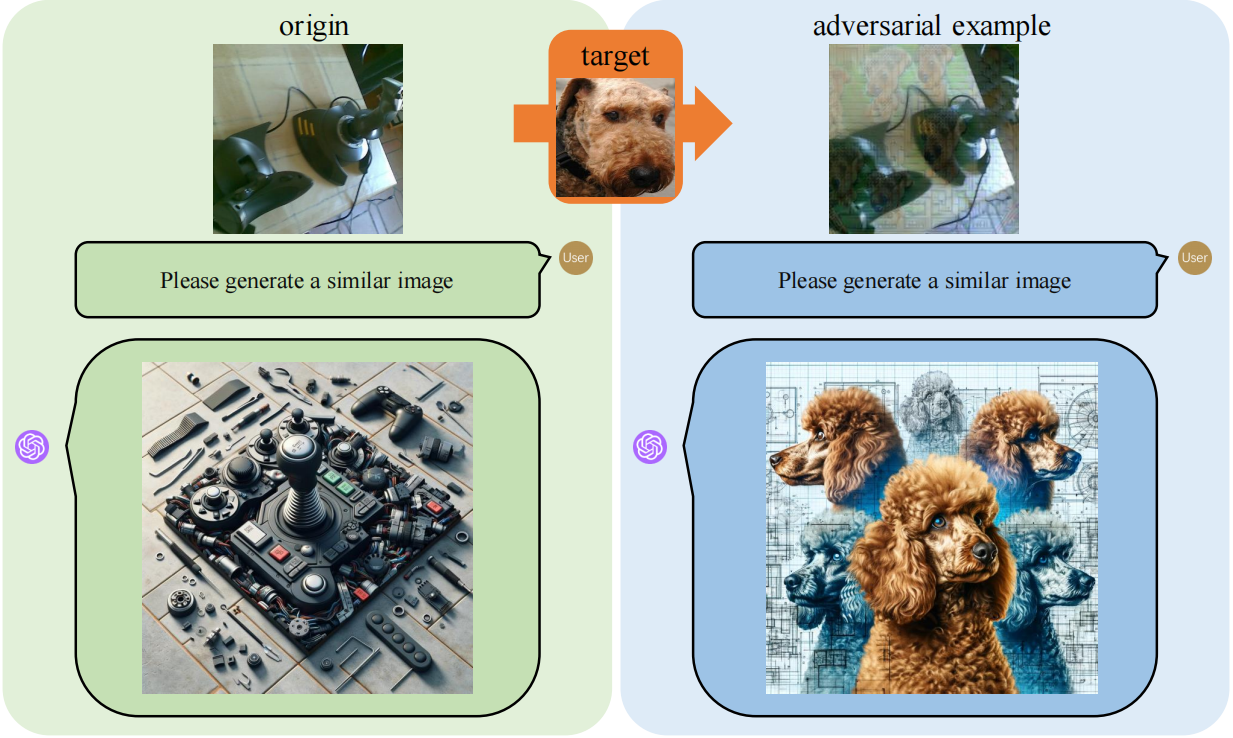}
  \caption{
From \textbf{"joystick"} to  \textbf{"dog"} generated by GPT-4V
}
  \label{fig:gpt0}
\end{figure}

When we ask GPT-4V to \textbf{"Please generate a similar image"}, as shown in \cref{fig:gpt0,fig:gpt1,fig:gpt2,fig:gpt3}, we can see that the generated sample is also similar to the target and utterly unrelated to the original class.

This is a strong indication that when dealing with Large Vision-Language Models, our method produces adversarial samples that effectively fool the model.

\section{Visualization of Adversarial Examples}

\begin{figure}[t]
  \centering
  \includegraphics[width=\linewidth]{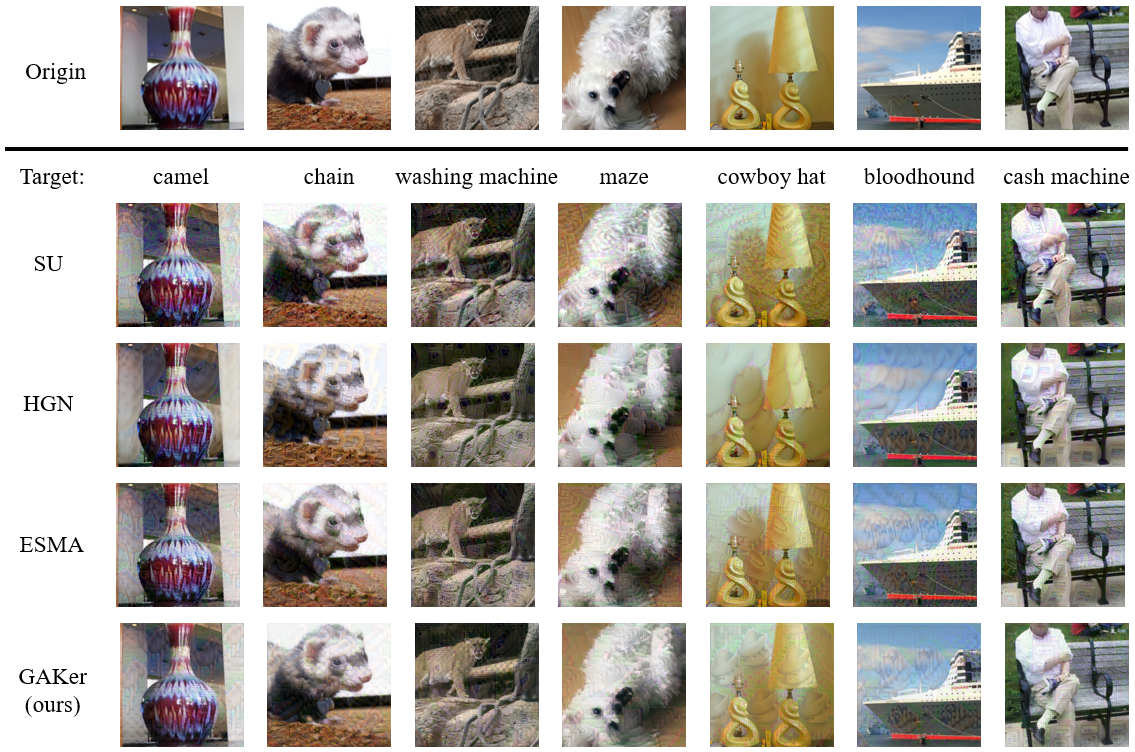}
  \caption{\textbf{Visualization of adversarial examples.} All adversarial examples are generated on Res-50.
}
  \label{fig:adv}
\end{figure}

In this section, we show some adversarial examples in \cref{fig:adv}.

\clearpage  

%
%

\end{document}